\documentclass[10pt,twocolumn,letterpaper]{article}

\usepackage{cvpr}
\usepackage{times}
\usepackage{epsfig}
\usepackage{graphicx}
\usepackage{amsmath}
\usepackage{amssymb}
\usepackage{amsthm}
\usepackage{xcolor}
\usepackage{xspace}
\usepackage{xfrac}
\usepackage{acronym}
\usepackage{bm}
\usepackage{mathtools}
\usepackage{enumitem}
\usepackage{tabularx}
\usepackage{booktabs}
\usepackage{float}
\usepackage{units}
\usepackage{algpseudocode}
\usepackage{algorithm}
\usepackage[numbers]{natbib}

\usepackage[utf8]{inputenc}
\usepackage[T1]{fontenc}
\DeclareUnicodeCharacter{221E}{$\infty$}

\makeatletter
\@namedef{ver@everyshi.sty}{}
\makeatother
\usepackage{tikz}
\usepackage{pgfplots}

\acrodef{pgd}[PGD]{projected gradient descent}

\renewcommand{\citet}[1]{\cite{#1}}

\newcommand{\algname}{Combined Abstention Robustness Learning\xspace}
\newcommand{\algac}{\mbox{CARL}\xspace}
\newcommand{\dfname}{DeepFool-Abstain\xspace}
\newcommand{\example}{\mathbf{x}}
\newcommand{\advexample}{\widetilde{\mathbf{x}}}
\newcommand{\clf}{f}
\newcommand{\baseclf}{g}
\newcommand{\baselineclf}{\clf_\text{baseline}}
\newcommand{\marginthresh}{{\gamma^{*}}}
\newcommand{\signmargin}{\hat{\gamma}}
\newcommand{\abstainclass}{a}
\newcommand{\indic}[1]{\textbf{1}\{#1\}}
\newcommand{\smallparagraph}[1]{\textbf{#1}\quad}

\newtheorem{theorem}{Theorem}
\newtheorem{lemma}{Lemma}

\AtBeginDocument{%
 \abovedisplayskip=6pt plus 2pt minus 4pt
 \abovedisplayshortskip=0pt plus 2pt
 \belowdisplayskip=6pt plus 2pt minus 4pt
 \belowdisplayshortskip=4pt plus 2pt minus 2pt
}

\usepackage[pagebackref=true,breaklinks=true,letterpaper=true,colorlinks,bookmarks=false]{hyperref}

\DeclareMathOperator{\sign}{sgn}
\DeclareMathOperator*{\argmin}{arg\,min}

\cvprfinalcopy % *** Uncomment this line for the final submission

\def\cvprPaperID{6748} % *** Enter the CVPR Paper ID here

\ifcvprfinal\fi
\begin{document}

%%%%%%%%% TITLE
\title{Playing it Safe: Adversarial Robustness with an Abstain Option}

\author{Cassidy Laidlaw\\
University of Maryland\\
College Park, MD\\
{\tt\small claidlaw@umd.edu}
\and
Soheil Feizi\\
University of Maryland\\
College Park, MD\\
{\tt\small sfeizi@cs.umd.edu}
}

\maketitle
%\thispagestyle{empty}

%%%%%%%%% ABSTRACT
\begin{abstract}
We explore adversarial robustness in the setting in which it is acceptable for a classifier to abstain---that is, output no class---on adversarial examples. Adversarial examples are small perturbations of normal inputs to a classifier that cause the classifier to give incorrect output; they present security and safety challenges for machine learning systems. In many safety-critical applications, it is less costly for a classifier to abstain on adversarial examples than to give incorrect output for them. We first introduce a novel objective function for adversarial robustness with an abstain option which characterizes an explicit tradeoff between robustness and accuracy. We then present a simple baseline in which an adversarially-trained classifier abstains on all inputs within a certain distance of the decision boundary, which we theoretically and experimentally evaluate. Finally, we propose \algname (\algac), a method for jointly learning a classifier and the region of the input space on which it should abstain. We explore different variations of the PGD and DeepFool adversarial attacks on \algac in the abstain setting. Evaluating against these attacks, we demonstrate that training with \algac results in a more accurate, robust, and efficient classifier than the baseline.
\end{abstract}

%%%%%%%%% BODY TEXT
\section{Introduction}

Many modern machine learning models are vulnerable to adversarial examples: slight perturbations of natural inputs that cause the model to behave erroneously \citep{szegedy_intriguing_2014}. For instance, stickers placed on a stop sign may fool an object detector to output that it is a speed limit sign instead \citep{eykholt_robust_2018}. Adversarial examples present a security challenge in safety-critical applications of machine learning, such as self-driving cars and medical diagnosis. Much research has focused on methods for ensuring that machine learning models give correct output, even when under adversarial attack \citep{goodfellow_explaining_2015,madry_towards_2018,zhang_theoretically_2019}.

However, for many applications, safety under adversarial attack may be achieved even if the model cannot determine the correct output, as long as it knows that it is uncertain about that output. That is, under adversarial attack, it may be acceptable for a model to give no output, or \textit{abstain}, for a particular input that it is uncertain about. For instance, if a medical diagnosis model believes it is under adversarial attack, it may give no output and instead tell the user that further tests are necessary. If a self-driving car detects an object that it is uncertain about---potentially an adversarial attack---it could avoid the object, come to a controlled stop, or ask a human to take over. All of these outcomes are much preferable to the model giving the \textit{wrong} answer. Furthermore, they do not require the model to be able to determine the right answer, only to determine that it is too uncertain to give one.

\begin{figure*}
    \centering
    \input{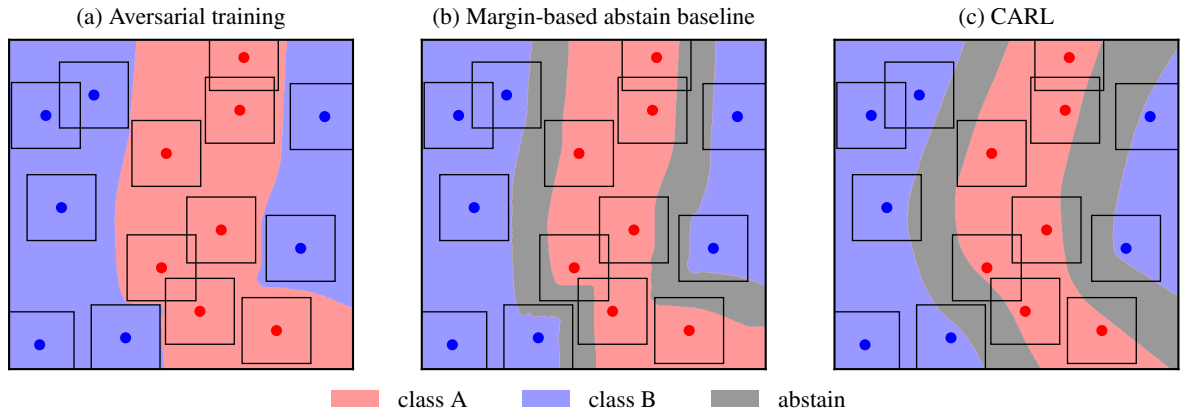}
    \caption{Classifiers trained on a 2D dataset with classes A (red) and B (blue). Training points are shown along with their $L_\infty$ neighborhood in which each classifier should be robust; we require that the classifier predicts correctly for natural inputs, but it may predict correctly \textit{or} abstain (predict neither class) at every point within the neighborhood. (a) Ordinary adversarial training is robust, but requires a complex decision boundary that ``weaves" between the two classes. (b) The baseline we explore in Section \ref{margin_baseline} abstains on inputs within a distance $\marginthresh$ of the decision boundary, but remains overly complex. (c) Using the proposed \algname (\algac), we jointly learn a classifier and abstain region, requiring a simpler decision boundary and reduced model capacity.}
    \label{fig:linf_2d_abstain}
\end{figure*}

Given this line of reasoning, we construct an alternate definition of adversarial robustness in which a classifier may abstain on certain inputs rather than determine an output class. We penalize the model for abstaining on natural, unperturbed inputs, but do not penalize it for abstaining on adversarial examples. This definition implies a tradeoff between natural error and adversarial error. On the one hand, the model may never abstain on an input, in which case it is likely to be accurate on natural data but vulnerable to adversarial attacks. On the other hand, the model may abstain on every input, which prevents adversarial vulnerability but is useless on natural inputs. We believe that ideal models for various applications exist between these extremes. See Figure \ref{fig:linf_2d_abstain} for an example of how a model can give correct predictions for natural inputs but abstain on inputs in adversarially vulnerable regions.

In Section \ref{margin_baseline}, we investigate a simple heuristic for a classifier to choose inputs to abstain on. We train a robust base classifier using adversarial training \cite{madry_towards_2018} and then abstain on any input which is within a particular distance $\marginthresh$ of a decision boundary of this base classifier. In this case, we describe the tradeoff between natural and adversarial error in Theorem \ref{thm:margin}, which gives a precise value for natural error and an upper bound for adversarial error given a base classifier and a margin threshold $\marginthresh$. 

This baseline has drawbacks. First, determining the distance to the decision boundary at inference time in a deep network requires solving a non-convex optimization, which is computationally expensive even to approximate. Second, the baseline is inflexible; we hypothesize that we may leverage the internal representations of the classifier to learn a better region over which to abstain.

To address these drawbacks, in Section \ref{learn_abstain}, we introduce a method for jointly learning the classifier and the areas of input space on which the classifier should abstain with a single neural network. We call this method \algname (\algac). The setting of adversarial robustness with an abstain option presents multiple optimization challenges for \algac. First, directly minimizing the adversarial and natural error is intractable. Thus, we propose surrogate loss functions for training that allow the classifier to either predict correctly or abstain on adversarial examples. Second, adversarial examples generated during training to enforce robustness must fool the classifier to both give an incorrect prediction \textit{and} not abstain. 

To solve the second problem, we develop six new variations of adversarial attacks based on \ac{pgd} \citep{madry_towards_2018} and DeepFool \citep{moosavi-dezfooli_deepfool:_2016} (see Section \ref{attacks_abstain}). These attacks all attempt to find adversarial examples that cause mis-classification outside of the abstain region. We use some attacks during training and then perform a comprehensive evaluation against the union of all six attacks in the white-box setting.

We train classifiers using the baseline method and \algac on MNIST and CIFAR-10. Both the baseline and \algac produce classifiers which can have arbitrarily low adversarial error at the cost of abstaining on some natural inputs. Furthermore, training with \algac gives lower natural and adversarial error than the baseline, showing the importance of jointly learning to classify and abstain. Our numerical results can be seen in Figure \ref{fig:error_plots} and Table \ref{tab:error_cifar}.

We hypothesize multiple explanations for why \algac outperforms the baseline. First, the baseline uses a single margin threshold for all inputs; it may instead be more useful to vary the margin threshold in various parts of the input space. Second, jointly training the classifier and abstain region with \algac can lead to simpler decision boundaries, since the decision boundary does not need to ``weave" between neighborhoods of training samples (see Figure \ref{fig:linf_2d_abstain}). Third, there are other reasons than proximity to a decision boundary to abstain on an input, such as the input being far from the training distribution. In fact, classifiers trained with \algac usually abstain on inputs of pure noise.

Our contributions are summarized as follows:

\begin{itemize}[leftmargin=*]
    \item We introduce a novel framework for adversarial robustness in which a classifier may abstain on some inputs rather than output a particular class. This framework is highly relevant for safety-critical applications in which the cost of an abstention is lower than that of an incorrect output.
    \item We explore a simple baseline for abstaining on inputs close to the decision boundary of a robust classifier and give a precise description of its accuracy-robustness tradeoff in Theorem \ref{thm:margin}.
    \item We propose \algname, a method for jointly learning a classifier with a region over which that classifier should abstain. We experimentally show that it outperforms the Pareto frontier of the already-strong baseline.
\end{itemize}

\section{Related Work}
\label{related_work}

\smallparagraph{Adversarial detection} The problem of abstaining on adversarial examples is closely related to  that of detecting adversarial examples, for which many solutions have been proposed. However, Carlini and Wagner \citep{carlini_adversarial_2017} have shown that many proposed detectors are vulnerable to attacks that attempt to fool both classifier and detector. Here, we only focus on detection approaches tested against these types of detector-aware attacks, since this is the type of attack we evaluate against. Pang et al. \citet{pang_towards_2018} attempt to detect adversarial examples far from the training distribution through an alternate training loss and kernel density estimation in representation space. Liang et al. \citet{liang_detecting_2018}'s detection method classifies an input image both before and after applying denoising and spatial smoothing; if the two classes output are not the same, they detect the input as an adversarial example. \citet{anonymous_adversarial_2019} begin with an already-trained classifier and train an additional detector for each class that attempts to distinguish adversarial examples from natural inputs. We believe that our method \algac outperforms these methods in robustness and efficiency; a more detailed comparison is in Appendix \ref{related_work_additional}.

\smallparagraph{Classification with an abstain option}
Some work has focused on training a classifier to abstain on certain inputs \citet{bartlett_classification_2008,herbei_classification_2006}.  However, little if any work has considered an abstain option in the context of adversarial robustness. While prior work has focused on adversarial detection, it has not explicitly described the motivation for detection and its resulting effect on the accuracy-robustness tradeoff.

\section{Problem Definition and Robustness Framework}

Our formulation of the problem of adversarial robustness is based around an observation that holds for many real-world applications: \textit{it is okay for a classifier to abstain---that is, output no class---when given an adversarial example}.
%This may be viewed equivalently as ``detecting" the attack. Understanding that a classification model is uncertain about its output is critical for making safety-conscious decisions.

\subsection{Tradeoff Between Robustness and Accuracy}

\begin{figure}
    \centering
    \input{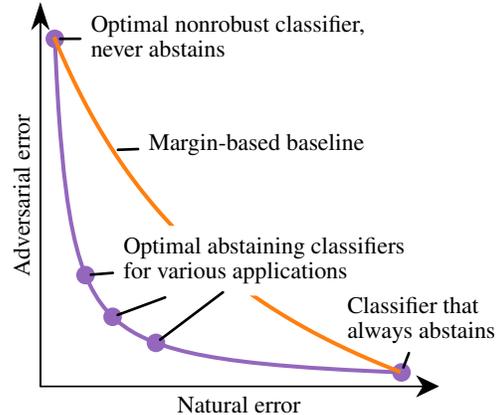}
    \caption{A conceptual illustration of the natural-adversarial error tradeoff for an abstaining classifier as defined in Section \ref{error_defs}. Always abstaining achieves 0\% adversarial error but has 100\% natural error. A nonrobust optimal classifier which never abstains likely achieves low natural error but may have high adversarial error. The baseline we describe in Section \ref{margin_baseline} can find classifiers between these extremes (orange line). However, we hypothesize that this baseline can be improved upon (purple line) to train abstaining classifiers with a more ideal accuracy-robustness tradeoff.}
    \label{fig:tradeoff}
\end{figure}

%Our problem definition, in which abstentions are not counted towards adversarial error,
This formulation has a simple solution: just abstain on every input. Since we would not like classifiers to do this, we also need to optimize for natural error: on natural, unperturbed inputs, a model should not abstain, and should instead output the correct class. Only on adversarial examples---perturbed inputs which are far from the training distribution---is the model allowed to abstain.

We thus propose a new framework for studying adversarial robustness with an abstain option. Under this framework, there are two types of error: natural error and adversarial error. Natural error is the proportion of naturally occurring inputs to a classifier which are misclassified or abstained on. Adversarial error is the proportion of naturally occurring inputs which can be perturbed to be both misclassified and not abstained on. We would like to minimize some combination of these two types of error.

As mentioned above, adversarial error alone can be minimized by simply abstaining on every input. Natural error alone can be minimized by using an optimal classifier with no regard for adversarial robustness and never abstaining. Most applications are likely best served by an approach between these two extremes; 
%The exact application affects the weighting of the two errors; for particularly safety-critical applications, one may accept more abstentions on natural examples if this allows lower adversarial error.
Figure \ref{fig:tradeoff} illustrates this tradeoff.

\subsection{Natural and Adversarial Error}
\label{error_defs}

Here, we formally define natural and adversarial error. Consider an input space $\mathbb{R}^d$ and set of labels $\mathcal{Y}$. Then an abstaining classifier $\clf(\cdot)$ is a function $\clf: \mathbb{R}^d \rightarrow \mathcal{Y} \cup \{\abstainclass\}$, which when given an input should either output a label in $\mathcal{Y}$ or the special class $\abstainclass$, which means the classifier has abstained on that input.

For an input $\example \in \mathbb{R}^d$, an adversarial example is a slight perturbation $\advexample$ of $\example$ such that $\clf(\advexample) \neq \clf(\example)$; that is, $\advexample$ is given a different label than $\example$ by the classifier. A slight perturbation is often defined as a small distance in some $L_p$ space, i.e. $\|\advexample - \example\|_p \leq \epsilon$, although other types of perturbations have been studied \citet{bhattad_big_2019,engstrom_rotation_2017,laidlaw_functional_2019,wong_wasserstein_2019,xiao_spatially_2018}. In this paper, we focus on the $L_\infty$ distance.

Denote by $\mathcal{D}$ the distribution of input-label pairs $(\example, y)$ and let $\indic{\text{event}}$ be the indicator function which is 1 if event occurs and 0 otherwise. Then the \textit{natural error} of $\clf(\cdot)$ is defined as the proportion of inputs which are classified incorrectly:
\begin{equation*}
    \mathcal{R}_\text{nat}(\clf) \triangleq \mathop{\mathbb{E}}_{(\example, y) \sim \mathcal{D}} \indic{f(\example) \neq y}
\end{equation*}
Let $\mathcal{B}_r (\example)$ be the open $L_\infty$ ball of radius $r$ around $\example$. The \textit{adversarial error} of $\clf(\cdot)$ is defined as the proportion of inputs for which an adversary can cause misclassification \textit{and} prevent the classifier from abstaining within such a ball:
\begin{equation}
    \label{eq:detect_risk}
    \mathcal{R}_\text{adv}(\clf) \triangleq \mathop{\mathbb{E}}_{(\example, y) \sim \mathcal{D}} \max_{\advexample \in \mathcal{B}_\epsilon (\example)} \indic{f(\advexample) \neq y \wedge f(\advexample) \neq \abstainclass}
\end{equation}
We propose optimizing a weighted sum of these two types of error, i.e. $\mathcal{R}(\clf) = \mathcal{R}_\text{nat}(\clf) + \lambda \, \mathcal{R}_\text{adv}(\clf)$. The term $\lambda$ in this formulation controls the tradeoff between natural and adversarial error. If $\lambda$ is low, then the model may allow more adversarial errors, but will usually give the correct class for a natural input rather than abstaining. If $\lambda$ is high, the resulting model will allow few adversarial errors, but may often abstain on natural inputs; this could be useful for safety-critical applications where avoiding failures is paramount and abstaining on an input is low-cost in comparison.

\section{A Margin-Based Abstention Baseline}
\label{margin_baseline}

We first investigate a simple baseline for robustly abstaining on adversarial examples. The baseline builds from a classifier $\baseclf (\cdot)$ which cannot abstain but is already robust to adversarial examples; this classifier could be trained by existing methods such as adversarial training \citep{madry_towards_2018}. Define $\gamma(\example)$ as the geometric margin (distance to a decision boundary) of $\example$ for the classifier $\baseclf (\cdot)$. That is,
\begin{equation}
    \gamma(\example) = \max r \;\text{s.t.}\; \forall \example' \in \mathcal{B}_r(\example), \baseclf(\example') = \baseclf(\example)
\end{equation}
The classifier $\baseclf (\cdot)$ is extended to one $\baselineclf (\cdot)$ that can additionally abstain as follows:
\begin{equation}
    \label{eq:baselineclf_def}
    \baselineclf (\example) =
    \begin{cases}
      \abstainclass \quad & \gamma(\example) \leq \marginthresh \\
      \baseclf(\example) & \text{otherwise}
    \end{cases}
\end{equation}
$\baselineclf (\cdot)$ will abstain on any inputs which are within a certain distance $\marginthresh \geq 0$ of a decision boundary of the classifier $\baseclf (\cdot)$. Intuitively, most natural examples should be far from a decision boundary in a robust classifier, and most adversarial examples should barely cross a decision boundary.

%Another interpretation is that predictions made on inputs close to the decision boundary are more uncertain, so it makes sense to abstain on these.

An issue with this formulation is that for neural networks it is generally infeasible to compute the geometric margin of an input. 
%Thus, how will we determine if $\gamma(\example) > \marginthresh$?
Instead of attempting to compute an exact margin, we determine whether to abstain by using an adversarial attack at inference time. Say we have a strong adversarial attack $\alpha_r: \mathbb{R}^d \times \mathcal{Y} \rightarrow \mathbb{R}^d$, which given an input example $\example$ and label $y$, attempts to find an adversarial example $\advexample = \alpha_r (\example, y)$ such that $\| \advexample - \example \| < r$. In general, if an adversarial attack succeeds it provides an upper bound on the geometric margin:
\begin{equation}
    \label{eq:attack_bounds_margin}
    \alpha_\marginthresh (\example, \baseclf(\example)) \neq \baseclf(\example)
    \quad \Rightarrow \quad
    \gamma(\example) \leq \marginthresh
\end{equation}
If the attack is strong---that is, it always finds an adversarial example if one exists---then the above becomes a bijection. Thus, we can apply to the input $\example$ an attack that attempts to change its label within the radius $\marginthresh$; if the attack succeeds, then the input's margin is less than $\marginthresh$. In this case, the classifier abstains rather than give the output of the base classifier. Formally,
\begin{equation*}
    \baselineclf (\example) =
    \begin{cases}
      \abstainclass \quad & \alpha_\marginthresh (\example, \baseclf(\example)) \neq \baseclf(\example) \\
      \baseclf(\example) & \text{otherwise}
    \end{cases}
\end{equation*}

\subsection{Robustness-Accuracy Tradeoff for Margin-Based Abstention}

\begin{figure}[t]
    \centering
    \input{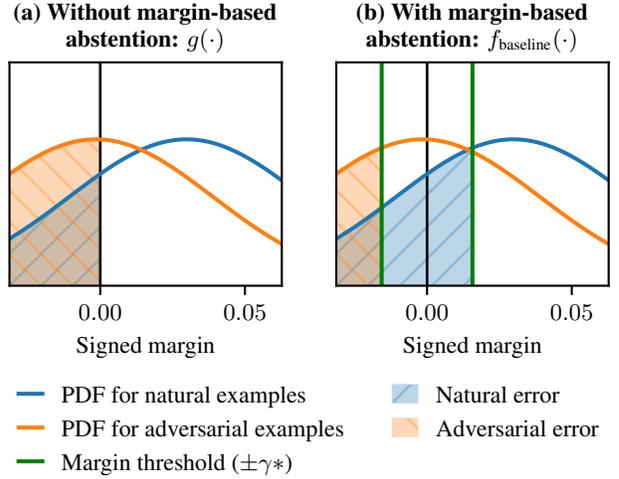}
    \caption{A visualization of the tradeoff between natural error and adversarial error described in Theorem \ref{thm:margin}. The PDFs of signed margins for natural and adversarial inputs to an adversarially-trained classifier on CIFAR-10 are shown. (a) With no abstain region, the classifier's natural error and adversarial error are the areas under the respective PDFs to the left of 0. (b) Adding an abstain region with threshold $\marginthresh$ (green lines) decreases adversarial error (orange region), since more adversarial examples are abstained on; however, it also increases natural error (blue region) due to false positives. For this classifier, a small value of $\marginthresh$ decreases adversarial error more than it increases natural error.}
    \label{fig:margin_dists}
\end{figure}

The threshold parameter $\marginthresh$ in the above formulation can be used to trade off natural error and adversarial error. Increasing $\marginthresh$ leads to lower adversarial error, because more adversarial examples are abstained on, but higher natural error, because some natural examples are mistakenly abstained on. Given the distribution of geometric margins, we can give a precise description of this tradeoff. Define the signed margin $\hat{\gamma}(\example, y)$ of $\example$ with respect to class $y$ as
\begin{equation*}
    \hat{\gamma} (\example, y) =
    \begin{cases}
      \gamma (\example) \quad & \baseclf(\example) = y \\
      -\gamma (\example) & \baseclf(\example) \neq y
    \end{cases}
\end{equation*}
Now let $\Gamma = \hat{\gamma}(\example, y), \; (\example, y) \sim \mathcal{D}$ be the distribution of signed margins for input-label pairs and let $F_{\Gamma}$ be its CDF, i.e. $F_{\Gamma}(\gamma) = \textbf{P} (\hat{\gamma}(\example, y) < \gamma \; | \; (\example, y) \sim \mathcal{D})$. Then we can state the following result about the natural and adversarial error of the derived classifier $\baselineclf (\cdot)$.

\begin{theorem}
\label{thm:margin}
Let $\baseclf: \mathbb{R}^d \rightarrow \mathcal{Y}$ be a classifier and let $F_\Gamma$ be the CDF of the distribution of signed margins for $\baseclf (\cdot)$. Define $\baselineclf: \mathbb{R}^d \rightarrow \mathcal{Y} \cup \{\abstainclass\}$ as in (\ref{eq:baselineclf_def}) with threshold parameter $\marginthresh$. Then for an attack radius $\epsilon$,
\begin{align*}
    \mathcal{R}_\text{nat}(\baselineclf) & = F_\Gamma(\marginthresh) \\
    \mathcal{R}_\text{adv}(\baselineclf) & \leq F_\Gamma(\epsilon - \marginthresh)
\end{align*}
\end{theorem}

The full proof can be found in Appendix \ref{margin_proof}. While calculating $F_\Gamma$ exactly is intractable for deep networks, it can be approximated using an adversarial attack as in (\ref{eq:attack_bounds_margin}); see also Appendix \ref{calculating_margin}.

\section{Jointly Learning a Robust Classifier and \hbox{Abstain} Region}
\label{learn_abstain}

The margin-based baseline for abstaining on adversarial examples described in Section \ref{margin_baseline} has two drawbacks. First, it has a fixed tradeoff between natural and adversarial error.
%; one can usually not reduce the adversarial error without increasing the natural error.
Second, it is computationally expensive to calculate, since an adversarial attack must be performed against every input at inference time. We propose a method to jointly learn a classifier and the region over which that classifier should abstain. This is in contrast to the baseline, in which a classifier is trained first and then an abstain region is calculated afterwards.

We propose to train a classifier $\clf: \mathbb{R}^d \rightarrow \mathcal{Y} \cup \{\abstainclass\}$ which learns to either output one of the classes $y \in \mathcal{Y}$ in the training data or an additional ``abstain" class $\abstainclass$. We would like $\clf(\cdot)$ to minimize a combination $\mathcal{R}(\clf) =  \mathcal{R}_\text{nat}(\clf) + \lambda\, \mathcal{R}_\text{adv}(\clf)$ of robust and adversarial error as defined in Section \ref{error_defs}. However, directly minimizing the 0-1 losses in these formulations is intractable, so instead we experiment with surrogate loss functions.

As a surrogate for natural error $\mathcal{R}_\text{nat}$, we use the standard cross-entropy loss. Let $p_i(\example)$ be the probability assigned by the classifier $\clf(\cdot)$ to class $i$ for input $\example$. Then the cross-entropy loss, also known as negative log likelihood loss, is given by
\begin{equation*}
    \mathcal{L}_\text{nat}(\clf) =
    \mathop{\mathbb{E}}_{(\example, y) \sim \mathcal{D}}
    -\log p_y(\example)
\end{equation*}

\begin{figure}[t]
    \centering
    \includegraphics[width=\columnwidth]{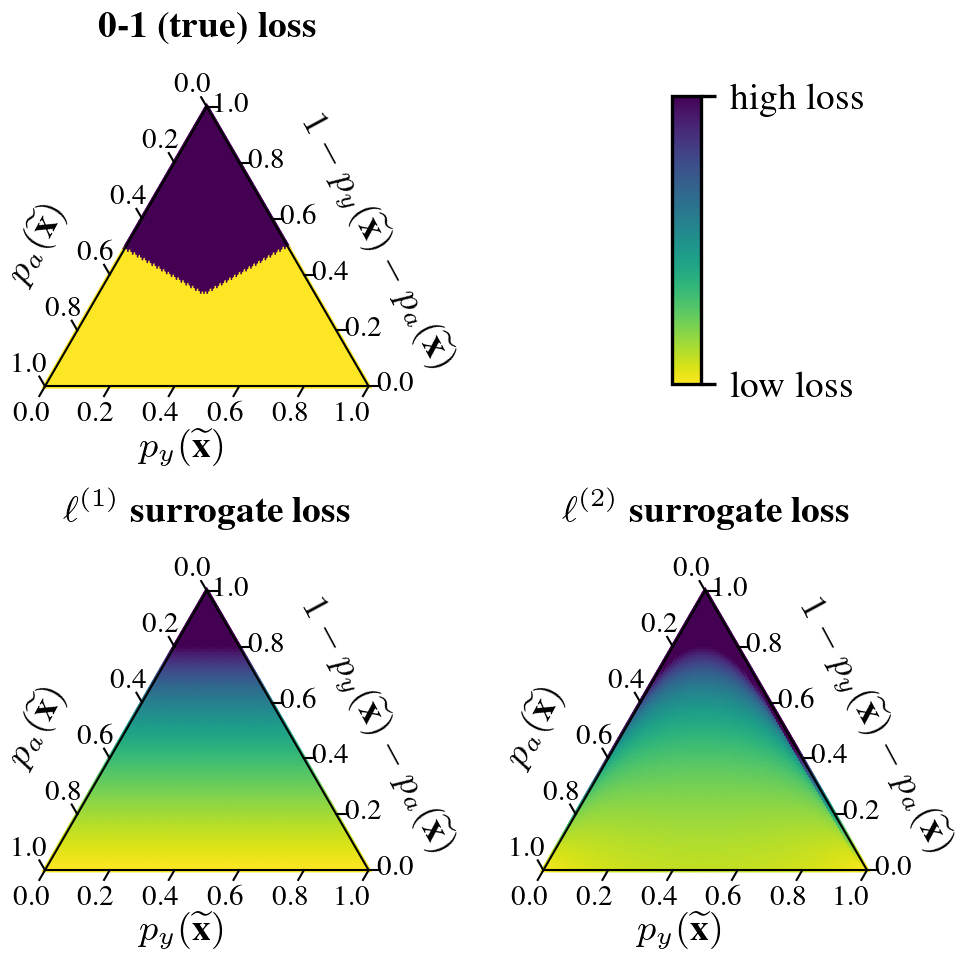}
    \caption{A comparison of the true 0-1 loss with surrogate losses from Section \ref{learn_abstain} for adversarial examples with an abstain option. The loss functions are shown as heatmaps in terms of the probability output for the correct class $p_y(\advexample)$ and for the abstain class $p_a(\advexample)$. Each loss is low as long as either the correct or abstain classes have high probability for an adversarial example.}
    \label{fig:ternary_losses}
\end{figure}

Often, in adversarial training, the adversarial loss is given by the maximum of some function $\ell(\clf, \advexample, y)$ over an $\epsilon$-ball around $\example$:
\begin{equation}
    \label{eq:training_loss_adv}
    \mathcal{L}_\text{adv}(\clf) =
    \mathop{\mathbb{E}}_{(\example, y) \sim \mathcal{D}}
    \max_{\advexample \in \mathcal{B}_\epsilon(\example)}
    \ell(\clf, \advexample, y)
\end{equation}
For instance, \citet{madry_towards_2018} use $\ell(\clf, \advexample, y) = -\log p_y(\advexample)$. In the formulation of robustness given in Section \ref{error_defs}, the classifier may output either the true class $y$ or the abstain class $\abstainclass$ within the $\epsilon$-ball. Thus, we would like an alternate formulation of $\ell(\clf, \advexample, y)$ which is low when either the correct class or the adversarial class is likely to be predicted.  In particular, here are some properties we would like to satisfy:
\begin{align*}
    \ell(\clf, \advexample, y) \rightarrow 0
    & \quad \text{as} \quad
    p_y(\advexample) \rightarrow 1 \\
    \ell(\clf, \advexample, y) \rightarrow 0
    & \quad \text{as} \quad
    p_\abstainclass(\advexample) \rightarrow 1 \\
    \ell(\clf, \advexample, y) \rightarrow \infty
    & \quad \text{as} \quad
    p_y(\advexample) + p_\abstainclass(\advexample) \rightarrow 0
\end{align*}
We experiment with two versions of $\ell(\clf, \advexample, y)$ which satisfy these properties:
\begin{align}
    \label{eq:adv_loss_inner_1}
    \ell^{(1)}(\clf, \advexample, y) & =
    -\log \left( p_y(\advexample) + p_\abstainclass(\advexample) \right) \\
    \label{eq:adv_loss_inner_2}
    \ell^{(2)}(\clf, \advexample, y) & =
    \big( -\log p_y(\advexample) \big)
    \big( -\log p_\abstainclass(\advexample) \big)
\end{align}
These surrogates are displayed in comparison with the true 0-1 loss in Figure \ref{fig:ternary_losses}. During training, they are used in the adversarial loss term (\ref{eq:training_loss_adv}).
% They give two formulations for the adversarial surrogate loss $\mathcal{L}_\text{adv}$:
% \begin{align*}
%     \mathcal{L}_\text{adv}^{(1)}(\clf, \advexample, y) & =
%     \mathop{\mathbb{E}}_{\mathcal{D}} \;
%     \max_{\advexample \in \mathcal{B}_\epsilon(\example)}
%     -\log \left( p_y(\advexample) + p_\abstainclass(\advexample) \right) \\
%     \mathcal{L}_\text{adv}^{(2)}(\clf, \advexample, y) & =
%     \mathop{\mathbb{E}}_{\mathcal{D}} \;
%     \max_{\advexample \in \mathcal{B}_\epsilon(\example)}
%     \big( -\log p_y(\advexample) \big)
%     \big( -\log p_\abstainclass(\advexample) \big)
% \end{align*}
In the next section, we present modifications of standard adversarial attacks that may be used to approximate the inner maximization of this formulation.

\section{Attacks on Abstaining Classifiers}
\label{attacks_abstain}

We develop multiple adaptations of the standard \ac{pgd} \citep{madry_towards_2018} and DeepFool \citep{moosavi-dezfooli_deepfool:_2016} adversarial attacks to the setting where an abstaining classifier $\clf (\cdot)$ must be fooled to both misclassify and not abstain. The attacks are useful for both solving the inner maximization in the adversarial training approach described in Section \ref{learn_abstain} as well as evaluating trained abstaining classifiers.

\subsection{Attacks Based on PGD}
\label{attacks_pgd}

\begin{figure}[t]
    \centering
    \input{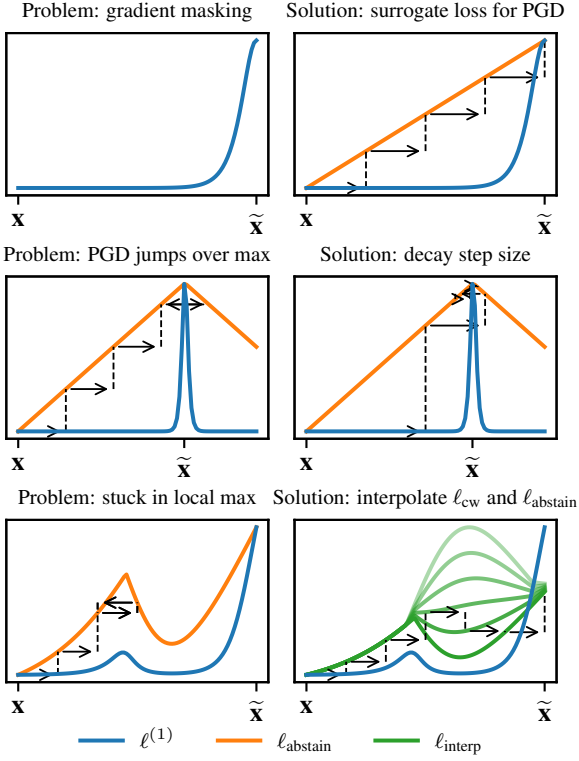}
    \caption{Various problems we solved when trying to find an adversarial example $\advexample$ during \algac training using PGD. \textbf{Top:} directly maximizing the adversarial loss $\ell^{(1)}$ or $\ell^{(2)}$ is prone to gradient masking, so instead we use $\ell_\text{abstain}$ or $\ell_\text{interp}$. \textbf{Middle:} PGD can sometimes ``jump" over sharp maxima, so we use an exponentially decaying step size. \textbf{Bottom:} PGD can get stuck in local maxima, so we vary the loss function from the beginning of PGD to the end.}
    \label{fig:pgd_fixes}
\end{figure}

\ac{pgd} is an iterative method for attacking a classifier $\clf (\cdot)$ given a loss function $\ell(f, \advexample, y)$ which is high when misclassification occurs. At each iteration $t$, the current input $\advexample^{(t)}$ is perturbed by step size $\delta$ in the direction $\nabla_{\advexample} \ell(f, \advexample^{(t)}, y)$. Then, the resulting input is projected into the ball of allowable perturbations $\mathcal{B}_\epsilon(\example)$. This process repeats for a set number of iterations or until misclassification is achieved.

For the standard adversarial robustness setting, a common choice of $\ell(f, \advexample, y)$ is given by Carlini and Wagner \citep{carlini_towards_2017}. Let $z_i(\example)$ be the $i$th logit (pre-softmax activation) of the classifier $\clf(\cdot)$ for input $\example$. Carlini and Wagner's loss is
\begin{equation*}
    \ell_\text{cw}(f, \advexample, y) =
    \max_{i \neq y} \big(z_i(\advexample)\big) - z_y(\advexample)
\end{equation*}
Maximizing this loss function reduces the probability the classifier outputs the correct class $y$ while increasing the probability the classifier outputs a different class. We can directly adapt this loss function to the case where our attack must both fool the classifier and prevent it from abstaining:
\begin{equation*}
    \ell_\text{abstain}(f, \advexample, y) =
    \max_{i \neq y, i \neq \abstainclass} \big(z_i(\advexample)\big) -
    \max \big(z_y(\advexample), z_\abstainclass(\advexample)\big)
\end{equation*}
We also experiment with three additional variations for $\ell_\text{abstain}(f, \advexample, y)$. Some of these variations depend on the current iteration $t$ out of $T$ total iterations of \ac{pgd}. They are
\begin{align*}
    \ell_\text{sum}(f, \advexample, y) & =
    \max_{i \neq y, i \neq \abstainclass} \big(z_i(\advexample)\big) - \big(z_y(\advexample) + z_\abstainclass(\advexample)\big) \\
    \ell_\text{interp}(f, \advexample, y) & =
    (1 - \sfrac{t}{T}) \ell_\text{cw}(f, \advexample, y) +
    (\sfrac{t}{T}) \ell_\text{abstain}(f, \advexample, y) \\
    \ell_\text{switch}(f, \advexample, y) & =
    \begin{cases}
      \ell_\text{cw}(f, \advexample, y) \quad & t < \sfrac{T}{2} \\
      \ell_\text{abstain}(f, \advexample, y) & \text{otherwise}
    \end{cases}
\end{align*}
$\ell_\text{sum}$ is similar to $\ell_\text{abstain}$ except that it sums $z_y(\advexample)$ and $z_\abstainclass(\advexample)$ instead of taking their maximum. $\ell_\text{interp}$ interpolates between $\ell_\text{cw}$ at the beginning of the attack and $\ell_\text{abstain}$ at the end. $\ell_\text{switch}$ uses $\ell_\text{cw}$ for half the iterations and then switches to $\ell_\text{abstain}$ for the remaining iterations. Both $\ell_\text{interp}$ and $\ell_\text{switch}$ first attack the true label $y$ and only later also attack the abstain class $\abstainclass$.

\subsection{Attacks Based on DeepFool}
\label{attacks_deepfool}

\begin{figure}[t]
    \centering
    \input{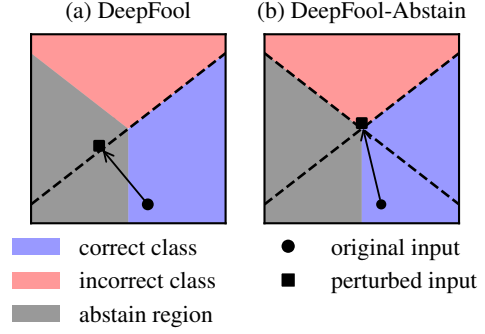}
    \caption{(a) Because DeepFool does not explicitly consider the abstain region, it may cross the linear approximation of the incorrect class's decision boundary but still remain unsuccessful. (b) \dfname (see Appendix \ref{deepfool_details}) considers the decision boundaries with both the correct class and abstain class, allowing it to reach the incorrect class in a single step.}
    \label{fig:deepfool_abstain}
\end{figure}

DeepFool \citep{moosavi-dezfooli_deepfool:_2016} is a generalized Newton's method \citep{atzmon_controlling_2019} for finding an adversarial example $\advexample$ close to an input $\example$. Instead of taking uniform step sizes as in \ac{pgd}, at each iteration $t$ DeepFool constructs a first-order approximation of the logits $z_i(\cdot)$ of the classifier at $\advexample^{(t)}$ and then calculates the smallest perturbation (in $L_p$ norm) to apply to $\advexample^{(t)}$ such that it will be misclassified under the approximation.

\begin{figure*}[!tb]
    \centering
    \input{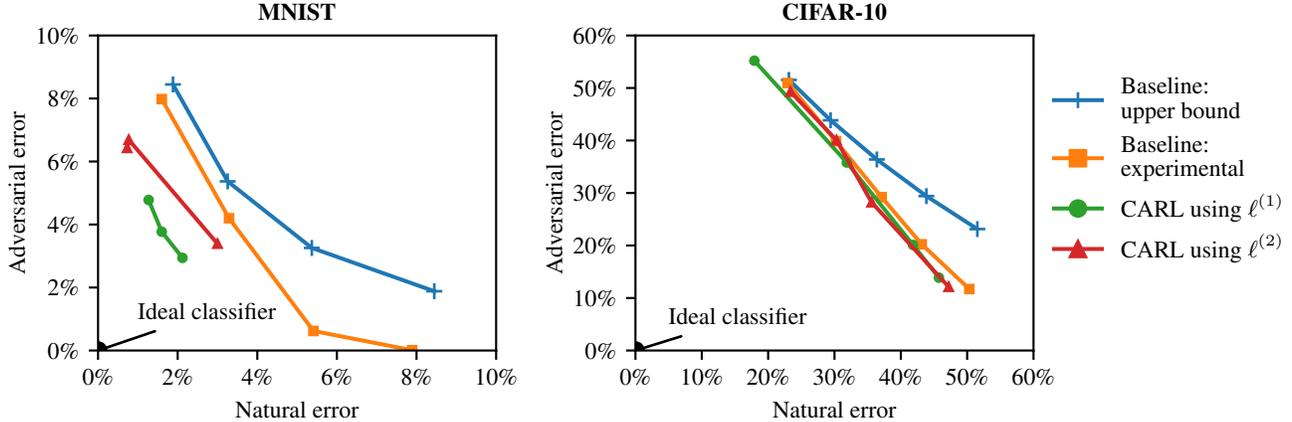}
    \caption{Natural and adversarial error, as defined in Section \ref{error_defs}, for the margin-based baseline as well as \algac-trained classifiers on MNIST and CIFAR-10; \algac improves on the baseline. Also, the bound given by Theorem \ref{thm:margin} (shown in blue) holds for the baseline.}
    \label{fig:error_plots}
\end{figure*}

We experiment with two variations of DeepFool applied to the setting where the classifier $f(\cdot)$ may abstain. In the first variation, we simply target all classes besides the correct and abstain class. In the second variation, which we call \dfname, a different approach is used to calculate the minimum perturbation for each incorrect class. Rather than only considering a linear approximation of the boundary between the incorrect class and the current class---which could be the correct class $y$ or the abstain class $\abstainclass$---it explicitly considers boundaries between the incorrect class $i$ and both $y$ and $\abstainclass$. For a conceptual diagram, see Figure \ref{fig:deepfool_abstain}. For the details of both DeepFool attacks, see Appendix \ref{deepfool_details}.

\section{Experiments}

\begin{table*}[t]
    \centering
    \begin{tabular}{lll|rrr|r|rrrrrrr}
        \toprule
        & & & \multicolumn{3}{c|}{\bf Natural Inputs} & \bf Noise & \multicolumn{7}{c}{\textbf{Adversarial Error}} \\
        \bf Defense &  &  & Acc. & Abs. & Inc. & Abs. & $\ell_\text{abstain}$ & $\ell_\text{sum}$ & $\ell_\text{interp}$ & $\ell_\text{switch}$ & DF & DF-Abs & Any \\
        \midrule
        \bf Baseline & \multicolumn{2}{l|}{$\marginthresh = 0$} & 77.1 & 0.0 & 22.9 & 0.0 & --- & --- & --- & --- & --- & --- & 50.9 \\
        \bf Baseline & \multicolumn{2}{l|}{$\marginthresh = \nicefrac{4}{255}$} & 62.9 & 30.1 & 7.0 & 94.0 & --- & --- & --- & --- & --- & --- & 29.2 \\
        \bf Baseline & \multicolumn{2}{l|}{$\marginthresh = \nicefrac{8}{255}$} & 49.7 & 47.7 & 2.6 & 100.0 & --- & --- & --- & --- & --- & --- & 11.7 \\
        \midrule
        \bf \algac & $\ell^{(1)}$ & $\lambda = \nicefrac{1}{2}$ & 82.1 & 3.7 & 14.2 & 0.0 & 53.2 & 40.3 & 53.7 & 53.9 & 54.0 & 52.6 & 55.2 \\
        \bf \algac & $\ell^{(1)}$ & $\lambda = 1$ & 68.1 & 30.0 & 1.8 & 100.0 & 27.1 & 13.5 & 28.7 & 30.1 & 25.8 & 23.6 & 35.8 \\
        \bf \algac & $\ell^{(1)}$ & $\lambda = 2$ & 58.2 & 41.3 & 0.5 & 99.7 & 15.0 & 10.9 & 16.0 & 15.8 & 13.1 & 12.2 & 20.1 \\
        \bf \algac & $\ell^{(1)}$ & $\lambda = 4$ & 54.3 & 45.3 & 0.4 & 100.0 & 11.3 & 8.0 & 11.0 & 10.7 & 7.2 & 6.8 & 13.9 \\
        \bf \algac & $\ell^{(2)}$ & $\lambda = \nicefrac{1}{4}$ & 76.6 & 16.8 & 6.6 & 100.0 & 45.1 & 26.9 & 42.2 & 45.2 & 48.1 & 45.8 & 49.3 \\
        \bf \algac & $\ell^{(2)}$ & $\lambda = \nicefrac{1}{3}$ & 69.7 & 27.3 & 3.0 & 100.0 & 29.3 & 15.1 & 33.1 & 35.2 & 35.9 & 33.1 & 40.1 \\
        \bf \algac & $\ell^{(2)}$ & $\lambda = \nicefrac{1}{2}$ & 64.4 & 34.3 & 1.3 & 99.9 & 18.5 & 10.2 & 21.4 & 22.3 & 15.7 & 13.7 & 28.2 \\
        \bf \algac & $\ell^{(2)}$ & $\lambda = 1$ & 52.8 & 46.5 & 0.8 & 100.0 & 7.7 & 4.9 & 9.0 & 7.6 & 4.3 & 4.1 & 12.1 \\
        \bottomrule
    \end{tabular}
    \vspace{6pt}
    \caption{CIFAR-10 evaluation of the margin-based abstention baseline and proposed method \algac for various loss function variations and values of hyperparameters $\marginthresh$ and $\lambda$. First shown is the percentage of natural inputs which are classified correctly, abstained on, and misclassified. The ``Noise" column contains the percentage of pure noise inputs which are abstained on. Finally, adversarial error is shown for PGD-100 (with the four proposed loss variants), the two DeepFool variants (with 50 iterations), and the union of all attacks.}
    \label{tab:error_cifar}
\end{table*}

We evaluate the baseline described in Section \ref{margin_baseline} and \algac on the MNIST \citep{lecun_mnist_2010} and CIFAR-10 \citep{krizhevsky_learning_2009} datasets. For MNIST, we train a 6-layer CNN and for CIFAR-10 we use a WideResNet-28-5 \citep{zagoruyko_wide_2016}, both with softmax activations, which we find to be as accurate as ReLU but less prone to gradient masking. All attacks use the $L_\infty$ norm with $\epsilon = 0.3$ for MNIST and $\epsilon = \nicefrac{8}{255}$ for CIFAR-10. Our full experimental setup is described in Appendix \ref{experimental_setup}.

\subsection{Baseline}

We first evaluate the baseline abstaining classifier $\baselineclf(\cdot)$ and experimentally verify Theorem \ref{thm:margin}. We train the robust base classifier $\baseclf(\cdot)$ with adversarial training using PGD \cite{madry_towards_2018}. 
%We estimate the distribution of signed margins $\Gamma$ for $\baseclf(\cdot)$ by attacking the classifier at different $L_\infty$ perturbation radiuses $\epsilon$.
Because a successful adversarial attack on an input bounds its geometric margin (see equation \ref{eq:attack_bounds_margin}), we can apply PGD to $\baseclf(\cdot)$ at a given radius $\epsilon$ to estimate the signed margin CDF $F_\Gamma(\epsilon)$. We fit a normal (Gaussian) CDF to the estimates at several $L_\infty$ radiuses and find that it is a good fit for small values of $\epsilon$ (see see Appendix \ref{calculating_margin} for details). Using this distribution and Theorem \ref{thm:margin}, we approximate the natural and adversarial error of $\clf(\cdot)$ for various margin thresholds; we use $\marginthresh = 0, 0.1, 0.2, 0.3$ for MNIST and $\marginthresh = 0, \sfrac{2}{255}, \sfrac{4}{255}, \sfrac{6}{255}, \sfrac{8}{255}$ for CIFAR-10. These theoretical upper bounds are shown as blue plus signs in Figure \ref{fig:error_plots}.

We also implement $\baselineclf(\cdot)$ as described in Section \ref{margin_baseline} by using an adversarial attack at inference time to determine if an input's geometric margin is less than the threshold $\marginthresh$. We attack $\baselineclf(\cdot)$ with PGD-100 and observe the resulting natural and adversarial error for the same values of $\marginthresh$; these results are shown as orange squares in Figure \ref{fig:error_plots}. The natural error predicted by Theorem \ref{thm:margin} is close to the natural error of $\baselineclf(\cdot)$ in practice; the adversarial error upper bound also holds, and is tight for small values of $\marginthresh$.

\subsection{\algac}

Next, we train an abstaining classifier using \algname (described in Section \ref{learn_abstain}) and the same hyperparameters as the baseline. Our training loss for an example is given by
\begin{equation*}
    \mathcal{L}(f, \example, y) =
    -\log p_y(\example) +
    \lambda \, \ell(f, \advexample, y) + \eta \left\| G_f(\example, y) \right\|_1
\end{equation*}
The first term is the standard cross-entropy loss. In the second term, $\ell$ is either $\ell^{(1)}$ or $\ell^{(2)}$ given in equations (\ref{eq:adv_loss_inner_1}-\ref{eq:adv_loss_inner_2}) and $\advexample \in \mathcal{B}_\epsilon(\example)$ is chosen via one of the attacks described below. The hyperparameter $\lambda$ controls a tradeoff between natural and adversarial error. In the third term,
\begin{equation*}
    G_f(\example, y) =
    \nabla_\example \big(z_y(\example) - \max_{i \neq y} z_i(\example) \big)
\end{equation*}
and thus this term attempts to regularize the $L_1$ norm of the Jacobian of the network. Others have explored Jacobian regularization to improve adversarial robustness \citep{jakubovitz_improving_2019,qin_adversarial_2019} and we find that it improves generalization in this setting.

\smallparagraph{Attacks during training}
We experimented with choosing $\advexample$ in the above loss function to directly maximize the second term. However, we find that the optimization is prone to gradient masking and often fails to converge. Instead, we choose $\advexample \in \mathcal{B}_\epsilon(\example)$ that maximizes one of the loss functions defined in Section \ref{attacks_pgd} by using \ac{pgd}; we find that attacks using $\ell_\text{abstain}$ and $\ell_\text{interp}$ work the best. In practice, a network trained against one of these attacks tends to be weak against the other, so during training we randomly sample a function from $\left\{ \ell_\text{abstain}, \ell_\text{interp} \right\}$ and use that to generate $\advexample$. For MNIST, we use $\ell_\text{sum}$ additionally.

We also find that the $\ell_\text{abstain}$ and $\ell_\text{sum}$ attack sometimes fails to find small ``gaps" between the correct and abstain classes; the step size during training is large enough that it jumps over the gap in a single step. Thus, for these attacks, we use an exponentially or harmonically decaying step size. See Appendix \ref{training_attacks_details} for full details and attack parameters; see Figure \ref{fig:pgd_fixes} for visualizations of how surrogate loss functions and decaying step size affect the search for $\advexample$.

\smallparagraph{Evaluation}
We evaluate our trained networks against six white-box attacks: four \ac{pgd}-based attacks using the loss functions in Section \ref{attacks_pgd} and the two DeepFool-based attacks described in Section \ref{attacks_deepfool}. The adversarial error and natural error of classifiers trained with various values of $\lambda$ are shown as green circles and red triangles in Figure \ref{fig:error_plots}. For adversarial error we use the union of all attacks, i.e. we give the proportion of inputs which can be perturbed to be both misclassified and not abstained on by \textit{any} of the six attacks.

Training with \algac consistently outperforms the baseline in accuracy and robustness. On MNIST, \algac reduces the adversarial error of the baseline by more than half while retaining equal natural error. On CIFAR-10, \algac produces classifiers that surpass the Pareto frontier of the baseline. For both datasets, using either the strong baseline or \algac, we can achieve arbitrarily low adversarial error at the cost of abstaining on a greater proportion of natural inputs. The choice of loss function $\ell^{(1)}$ or $\ell^{(2)}$ during training does not seem to make a significant difference to the resulting classifiers' error rates. As expected, $\lambda$ allows control over the accuracy-robustness tradeoff.

\smallparagraph{Abstention as an indicator of uncertainty}
\algac only optimizes the trained classifier to abstain on adversarial examples. However, we find that classifiers trained with \algac use the abstain class to indicate uncertainty in other settings. First, these classifiers usually abstain on natural inputs when they cannot output the correct class. For instance, the \algac-trained classifier on CIFAR-10 with $\ell^{(1)}$ and $\lambda = 1$ outputs the correct class 68.2\% of the time, but it abstains on 30\% of inputs as well, meaning that it only gives an entirely incorrect prediction for 1.8\% of natural inputs. This is valuable for safety critical applications because it means the classifier rarely makes mistakes without understanding that it is uncertain.

Second, classifiers trained with \algac tend to abstain on inputs  of pure noise (see the ``Noise" column in tables \ref{tab:error_cifar} and \ref{tab:error_mnist}). Abstaining on noise indicates that such a classifier expresses uncertainty about inputs which are far from the training distribution. Hein et al. \citep{hein_why_2019} observe that neural networks tend to exhibit high confidence on noise inputs; this problem is largely ameliorated by training with \algac.

% \begin{figure}
%     \centering
%     \input{images/error_plot_cifar.pgf}
%     \caption{Natural and adversarial error, as defined in Section \ref{error_defs}, for various classifiers on CIFAR-10. Our learned abstaining classifier improves on the margin-threshold baseline.}
%     \label{fig:error_plot_cifar}
% \end{figure}

\section{Conclusion}
\label{conclusion}

We have introduced the framework of adversarial robustness with an abstain option and explored approaches for a classifier to robustly abstain on adversarial examples. The baseline approach we present has a straightfoward accuracy-robustness tradeoff. Our method \algac improves on this tradeoff and is much more efficient at inference time. Furthermore, it produces classifiers that abstain on uncertain natural inputs and points far from the training distribution. All these properties make \algac useful for safety-critical applications in which adversarial robustness is needed. We believe that classification with an abstain option presents a promising direction for further research into mitigating adversarial attacks; Shafahi et al. \citet{shafahi_are_2019} note that using a ``don't know" class may help to overcome fundamental bounds on robustness. In the future, we hope to further improve the accuracy-robustness tradeoff of this formulation, moving towards safe machine learning models which are robust and uncertainty-aware.

{\small
\bibliographystyle{ieee_fullname}
\bibliography{paper}

\begin{thebibliography}{10}\itemsep=-1pt

\bibitem{anonymous_adversarial_2019}
Anonymous.
\newblock Adversarial {Example} {Detection} and {Classification} with
  {Asymmetrical} {Adversarial} {Training}.
\newblock {\em Under review for ICLR 2020}, Sept. 2019.

\bibitem{atzmon_controlling_2019}
Matan Atzmon, Niv Haim, Lior Yariv, Ofer Israelov, Haggai Maron, and Yaron
  Lipman.
\newblock Controlling {Neural} {Level} {Sets}.
\newblock {\em arXiv:1905.11911 [cs, stat]}, Oct. 2019.
\newblock arXiv: 1905.11911.

\bibitem{bartlett_classification_2008}
Peter~L. Bartlett and Marten~H. Wegkamp.
\newblock Classification with a {Reject} {Option} {Using} a {Hinge} {Loss}.
\newblock {\em Journal of Machine Learning Research}, 9(Aug):1823--1840, 2008.

\bibitem{bhattad_big_2019}
Anand Bhattad, Min~Jin Chong, Kaizhao Liang, Bo Li, and David~A. Forsyth.
\newblock Big but {Imperceptible} {Adversarial} {Perturbations} via {Semantic}
  {Manipulation}.
\newblock {\em arXiv preprint arXiv:1904.06347}, 2019.

\bibitem{cadzow_finite_1973}
James~A. Cadzow.
\newblock A {Finite} {Algorithm} for the {Minimum} {L}∞ {Solution} to a
  {System} of {Consistent} {Linear} {Equations}.
\newblock {\em SIAM Journal on Numerical Analysis}, 10(4):607--617, 1973.

\bibitem{carlini_adversarial_2017}
Nicholas Carlini and David Wagner.
\newblock Adversarial {Examples} are {Not} {Easily} {Detected}: {Bypassing}
  {Ten} {Detection} {Methods}.
\newblock In {\em Proceedings of the 10th {ACM} {Workshop} on {Artificial}
  {Intelligence} and {Security}}, pages 3--14. ACM, 2017.

\bibitem{carlini_towards_2017}
Nicholas Carlini and David Wagner.
\newblock Towards {Evaluating} the {Robustness} of {Neural} {Networks}.
\newblock In {\em 2017 {IEEE} {Symposium} on {Security} and {Privacy} ({SP})},
  pages 39--57. IEEE, 2017.

\bibitem{engstrom_rotation_2017}
Logan Engstrom, Brandon Tran, Dimitris Tsipras, Ludwig Schmidt, and Aleksander
  Madry.
\newblock A {Rotation} and a {Translation} {Suffice}: {Fooling} {CNNs} with
  {Simple} {Transformations}.
\newblock {\em arXiv preprint arXiv:1712.02779}, 2017.

\bibitem{eykholt_robust_2018}
Kevin Eykholt, Ivan Evtimov, Earlence Fernandes, Bo Li, Amir Rahmati, Chaowei
  Xiao, Atul Prakash, Tadayoshi Kohno, and Dawn Song.
\newblock Robust {Physical}-{World} {Attacks} on {Deep} {Learning} {Models}.
\newblock In {\em {CVPR} 2018}, Apr. 2018.
\newblock arXiv: 1707.08945.

\bibitem{goodfellow_explaining_2015}
Ian Goodfellow, Jonathon Shlens, and Christian Szegedy.
\newblock Explaining and {Harnessing} {Adversarial} {Examples}.
\newblock In {\em International {Conference} on {Learning} {Representations}},
  2015.

\bibitem{hein_why_2019}
Matthias Hein, Maksym Andriushchenko, and Julian Bitterwolf.
\newblock Why {ReLU} networks yield high-confidence predictions far away from
  the training data and how to mitigate the problem.
\newblock In {\em {CVPR}}, May 2019.
\newblock arXiv: 1812.05720.

\bibitem{herbei_classification_2006}
Radu Herbei and Marten~H. Wegkamp.
\newblock Classification with {Reject} {Option}.
\newblock {\em Canadian Journal of Statistics}, 34(4):709--721, 2006.

\bibitem{jakubovitz_improving_2019}
Daniel Jakubovitz and Raja Giryes.
\newblock Improving {DNN} {Robustness} to {Adversarial} {Attacks} using
  {Jacobian} {Regularization}.
\newblock {\em arXiv:1803.08680 [cs, stat]}, May 2019.
\newblock arXiv: 1803.08680.

\bibitem{kingma_adam:_2014}
Diederik~P. Kingma and Jimmy Ba.
\newblock Adam: {A} {Method} for {Stochastic} {Optimization}.
\newblock {\em arXiv preprint arXiv:1412.6980}, 2014.

\bibitem{krizhevsky_learning_2009}
Alex Krizhevsky and Geoffrey Hinton.
\newblock Learning {Multiple} {Layers} of {Features} from {Tiny} {Images}.
\newblock Technical report, Citeseer, 2009.

\bibitem{laidlaw_functional_2019}
Cassidy Laidlaw and Soheil Feizi.
\newblock Functional {Adversarial} {Attacks}.
\newblock {\em arXiv:1906.00001 [cs]}, Oct. 2019.
\newblock arXiv: 1906.00001.

\bibitem{lecun_mnist_2010}
Yann LeCun, Corinna Cortes, and CJ Burges.
\newblock {MNIST} {Handwritten} {Digit} {Database}.
\newblock {\em ATT Labs [Online]. Available: http://yann. lecun.
  com/exdb/mnist}, 2, 2010.

\bibitem{liang_detecting_2018}
Bin Liang, Hongcheng Li, Miaoqiang Su, Xirong Li, Wenchang Shi, and XiaoFeng
  Wang.
\newblock Detecting {Adversarial} {Image} {Examples} in {Deep} {Neural}
  {Networks} with {Adaptive} {Noise} {Reduction}.
\newblock {\em IEEE Transactions on Dependable and Secure Computing}, pages
  1--1, 2018.

\bibitem{madry_towards_2018}
Aleksander Madry, Aleksandar Makelov, Ludwig Schmidt, Dimitris Tsipras, and
  Adrian Vladu.
\newblock Towards {Deep} {Learning} {Models} {Resistant} to {Adversarial}
  {Attacks}.
\newblock In {\em International {Conference} on {Learning} {Representations}},
  2018.

\bibitem{moosavi-dezfooli_deepfool:_2016}
Seyed-Mohsen Moosavi-Dezfooli, Alhussein Fawzi, and Pascal Frossard.
\newblock Deepfool: a {Simple} and {Accurate} {Method} to {Fool} {Deep}
  {Neural} {Networks}.
\newblock In {\em Proceedings of the {IEEE} {Conference} on {Computer} {Vision}
  and {Pattern} {Recognition}}, pages 2574--2582, 2016.

\bibitem{pang_towards_2018}
Tianyu Pang, Chao Du, Yinpeng Dong, and Jun Zhu.
\newblock Towards {Robust} {Detection} of {Adversarial} {Examples}.
\newblock In S. Bengio, H. Wallach, H. Larochelle, K. Grauman, N. Cesa-Bianchi,
  and R. Garnett, editors, {\em Advances in {Neural} {Information} {Processing}
  {Systems} 31}, pages 4579--4589. Curran Associates, Inc., 2018.

\bibitem{paszke_automatic_2017}
Adam Paszke, Sam Gross, Soumith Chintala, Gregory Chanan, Edward Yang, Zachary
  DeVito, Zeming Lin, Alban Desmaison, Luca Antiga, and Adam Lerer.
\newblock Automatic {Differentiation} in {PyTorch}.
\newblock In {\em {NIPS}-{W}}, 2017.

\bibitem{qin_adversarial_2019}
Chongli Qin, James Martens, Sven Gowal, Dilip Krishnan, Krishnamurthy
  Dvijotham, Alhussein Fawzi, Soham De, Robert Stanforth, and Pushmeet Kohli.
\newblock Adversarial {Robustness} through {Local} {Linearization}.
\newblock {\em arXiv:1907.02610}, July 2019.
\newblock arXiv: 1907.02610.

\bibitem{shafahi_are_2019}
Ali Shafahi, W.~Ronny Huang, Christoph Studer, Soheil Feizi, and Tom Goldstein.
\newblock Are {Adversarial} {Examples} {Inevitable}?
\newblock In {\em {ICLR}}, 2019.

\bibitem{szegedy_intriguing_2014}
Christian Szegedy, Wojciech Zaremba, Ilya Sutskever, Joan Bruna, Dumitru Erhan,
  Ian Goodfellow, and Rob Fergus.
\newblock Intriguing {Properties} of {Neural} {Networks}.
\newblock In {\em International {Conference} on {Learning} {Representations}},
  2014.

\bibitem{wong_wasserstein_2019}
Eric Wong, Frank~R. Schmidt, and J.~Zico Kolter.
\newblock Wasserstein {Adversarial} {Examples} via {Projected} {Sinkhorn}
  {Iterations}.
\newblock {\em arXiv preprint arXiv:1902.07906}, Feb. 2019.
\newblock arXiv: 1902.07906.

\bibitem{xiao_spatially_2018}
Chaowei Xiao, Jun-Yan Zhu, Bo Li, Warren He, Mingyan Liu, and Dawn Song.
\newblock Spatially {Transformed} {Adversarial} {Examples}.
\newblock {\em arXiv preprint arXiv:1801.02612}, 2018.

\bibitem{zagoruyko_wide_2016}
Sergey Zagoruyko and Nikos Komodakis.
\newblock Wide {Residual} {Networks}.
\newblock In {\em British {Machine} {Vision} {Conference}}. British Machine
  Vision Association, 2016.

\bibitem{zhang_theoretically_2019}
Hongyang Zhang, Yaodong Yu, Jiantao Jiao, Eric~P. Xing, Laurent~El Ghaoui, and
  Michael~I. Jordan.
\newblock Theoretically {Principled} {Trade}-off {Between} {Robustness} and
  {Accuracy}.
\newblock In {\em {ICML}}, 2019.

\end{thebibliography}
}

\clearpage
\appendix

\section{Proof of Theorem \ref{thm:margin}}
\label{margin_proof}

Here we provide a full proof of Theorem \ref{thm:margin}. Throughout the proof, let $\baseclf: \mathbb{R}^d \rightarrow \mathcal{Y}$ be a classifier, and let $\gamma (\example)$ and $\signmargin (\example, y)$ denote the geometric margin and signed margin, respectively, of $\baseclf (\cdot)$ for $\example$. Define $\baselineclf: \mathbb{R}^d \rightarrow \mathcal{Y} \cup \{\abstainclass\}$ with threshold parameter $\marginthresh$ as

\begin{equation*}
    \baselineclf (\example) =
    \begin{cases}
      \abstainclass \quad & \gamma(\example) \leq \marginthresh \\
      \baseclf(\example) & \text{otherwise}
    \end{cases}
\end{equation*}

\begin{lemma}
\label{lemma:margin_ineq}
Let $\example_1, \example_2 \in \mathbb{R}^d$ and $y \in \mathcal{Y}$. Then
\begin{equation*}
    | \signmargin(\example_1, y) - \signmargin(\example_2, y) | \leq
    \| \example_1 - \example_2 \|
\end{equation*}
\end{lemma}
\begin{proof}
Assume that $\signmargin(\example_1, y) \neq \signmargin(\example_2, y)$; if $\signmargin(\example_1, y) = \signmargin(\example_2, y)$ then $0 \leq \| \example_1 - \example_2 \|$ is obviously true. We consider the cases when $\baseclf(\example_1) = \baseclf(\example_2)$ and $\baseclf(\example_1) \neq \baseclf(\example_2)$; in each case the proof is by contradiction.

\smallparagraph{Case 1} Say $\baseclf(\example_1) = \baseclf(\example_2)$. Note this implies that 
\begin{equation}
    \label{eq:signs_product}
    \sign(\signmargin(\example_1, y)) \sign(\signmargin(\example_2, y)) \geq 0
\end{equation}
%Then $\signmargin(\example_1, y) > \signmargin(\example_2, y) \geq 0$, $\signmargin(\example_1, y) = \gamma(\example_1)$, and $\signmargin(\example_2, y) = \gamma(\example_2)$.
Assume the lemma is false, that is,
\begin{equation}
    \label{eq:assume_not}
    \| \example_1 - \example_2 \| < | \signmargin(\example_1, y) - \signmargin(\example_2, y) |
\end{equation}
Without loss of generality, we may also assume that
\begin{equation}
    \label{eq:wlog}
    \gamma(\example_1) = |\signmargin(\example_1, y)| > |\signmargin(\example_2, y)| = \gamma(\example_2)
\end{equation}
Combining (\ref{eq:signs_product}), (\ref{eq:assume_not}), and (\ref{eq:wlog}), we have
$$\| \example_1 - \example_2 \| < | \signmargin(\example_1, y) - \signmargin(\example_2, y) |$$
$$= \Big| |\signmargin(\example_1, y)| - |\signmargin(\example_2, y)| \Big| = \gamma(\example_1) - \gamma(\example_2)$$
Then define
\begin{equation}
    \label{eq:rstar_1}
    r^* = \gamma(\example_1) - \| \example_1 - \example_2 \| > \gamma(\example_2)
\end{equation}
Let $\example'$ such that $\| \example' - \example_2 \| < r^*$. Then
$$\| \example' - \example_2 \| < \gamma(\example_1) - \| \example_1 - \example_2 \|$$
$$\| \example' - \example_2 \| + \| \example_2 - \example_1 \| < \gamma(\example_1)$$
$$\| \example' - \example_1 \| < \gamma(\example_1)$$
So by definition of geometric margin, $\baseclf(\example') = \baseclf(\example_1) = \baseclf(\example_2)$. Thus, we have
$$\| \example' - \example_2 \| < r^* \quad \Rightarrow \quad \baseclf(\example') = \baseclf(\example_2)$$
This implies $\gamma(\example_2) \geq r^*$, which is a contradiction since by (\ref{eq:rstar_1}) we have $\gamma(\example_2) < r^*$.

\smallparagraph{Case 2} Say $\baseclf(\example_1)  \neq \baseclf(\example_2)$. Again, assume the lemma is false; that is,
$$\| \example_1 - \example_2 \| < | \signmargin(\example_1, y) - \signmargin(\example_2, y) |$$
$$\leq | \signmargin(\example_1, y) | + | - \signmargin(\example_2, y) | = \gamma(\example_1) + \gamma(\example_2)$$
Since $\gamma(\example_1) + \gamma(\example_2) > 0$, this implies that
\begin{equation}
    \label{eq:margin_ratio}
    \frac{\| \example_1 - \example_2 \|}{\gamma(\example_1) + \gamma(\example_2)} < 1
\end{equation}
Now define
$$\example' = \frac{\example_1 \gamma(\example_2) + \example_2 \gamma(\example_1)}{\gamma(\example_1) + \gamma(\example_2)}$$
Calculate $\|\example' - \example_1\|$ as follows:
$$\|\example' - \example_1\| = \left\|\frac{\example_1 \gamma(\example_2) + \example_2 \gamma(\example_1)}{\gamma(\example_1) + \gamma(\example_2)} - \example_1\right\|$$
$$ = \left\|\frac{\example_1 \gamma(\example_2) + \example_2 \gamma(\example_1) - \example_1 \gamma(\example_1) - \example_1 \gamma(\example_2)}{\gamma(\example_1) + \gamma(\example_2)}\right\|$$
$$ = \left\|(\example_2 - \example_1) \frac{\gamma(\example_1)}{\gamma(\example_1) + \gamma(\example_2)}\right\|$$
$$ = \frac{\| \example_1 - \example_2 \|}{\gamma(\example_1) + \gamma(\example_2)} \gamma(\example_1) < \gamma(\example_1)$$
The last step is due to (\ref{eq:margin_ratio}). Since $\|\example' - \example_1\| < \gamma(\example_1)$,
$$\baseclf (\example') = \baseclf(\example_1)$$
However, a similar calculation shows that $\|\example' - \example_2\| < \gamma(\example_2)$ as well, so 
$$\baseclf (\example') = \baseclf(\example_2)$$
Thus, we have
$$\baseclf (\example') = \baseclf(\example_1) \neq \baseclf(\example_2) = \baseclf (\example')$$
which is a contradiction.
\end{proof}

Given lemma \ref{lemma:margin_ineq}, we can prove Theorem \ref{thm:margin}.

\addtocounter{theorem}{-1}
\begin{theorem}[restated]
Let $F_\Gamma$ be the CDF of the distribution of signed margins for $\baseclf (\cdot)$. Then for an attack radius $\epsilon$,

\begin{equation}
    \label{eq:margin_nat_risk}
    \mathcal{R}_\text{nat}(\baselineclf) = F_\Gamma(\marginthresh)
\end{equation}
\begin{equation}
    \label{eq:margin_adv_risk}
    \mathcal{R}_\text{adv}(\baselineclf) \leq F_\Gamma(\epsilon - \marginthresh)
\end{equation}
\end{theorem}
\begin{proof}
Note that for any event,
$$\mathbb{E} (\indic{\text{event}}) = \textbf{P} (\text{event})$$

\smallparagraph{Proof of (\ref{eq:margin_nat_risk})}
We would like to prove that
$$\mathop{\mathbb{E}}_{(\example, y) \sim \mathcal{D}} \indic{f(\example) \neq y} =
\textbf{P}(f(\example) \neq y) =
\textbf{P}(\signmargin(\example, y) \leq \marginthresh)$$
It suffices to show that
$$\signmargin(\example, y) \leq \marginthresh \quad \iff \quad f(\example) \neq y$$

First, we prove that $\signmargin(\example, y) \leq \marginthresh \; \Rightarrow \; f(\example) \neq y$. If $\signmargin(\example, y) \leq \marginthresh$, then either $\gamma(\example) = |\signmargin(\example, y)| \leq \marginthresh$ or $\signmargin(\example, y) < 0$. If $\gamma(\example) \leq \marginthresh$, then $\baselineclf(\example) = \abstainclass \neq y$ by definition of $\baselineclf (\cdot)$. If $\signmargin(\example, y) < 0$, then $\baseclf(\example) \neq y$. Either $\baselineclf(\example) = \baseclf(\example) \neq y$ or $\baselineclf(\example) = \abstainclass \neq y$, so $\baselineclf(\example) \neq y$.

Second, we prove the inverse: $\signmargin(\example, y) > \marginthresh \; \Rightarrow \; f(\example) = y$. Since $\signmargin(\example, y) > \marginthresh > 0$, $\baseclf(\example) = y$. Also, because $\signmargin(\example, y) = \gamma(\example) > \marginthresh$, $\baselineclf(\example) = \baseclf(\example)$. So $\baselineclf(\example) = \baseclf(\example) = y$.

\smallparagraph{Proof of (\ref{eq:margin_adv_risk})}
Similarly to the case of natural error, for adversarial error we would like to prove that
$$\mathop{\mathbb{E}}_{(\example, y) \sim \mathcal{D}} \max_{\advexample \in \mathcal{B}_\epsilon (\example)} \indic{f(\advexample) \neq y \wedge f(\advexample) \neq \abstainclass} =$$
$$\textbf{P}(\exists \advexample \in \mathcal{B}_\epsilon (\example) \; \text{s.t.} f(\advexample) \neq y \wedge f(\advexample) \neq \abstainclass) \leq \textbf{P}(\signmargin(\example, y) < \epsilon - \marginthresh)$$
In this case, because of the inequality, we only need to prove that
\begin{equation*}
\begin{split}
    \exists \advexample \in \mathcal{B}_\epsilon (\example) \; \text{s.t.} \baselineclf(\advexample) \neq y \wedge \baselineclf(\advexample) \neq \abstainclass \\
    \Rightarrow \quad
    \signmargin(\example, y) < \epsilon - \marginthresh
\end{split}
\end{equation*}
In fact, we prove the contrapositive:
\begin{equation*}
\begin{split}
    \signmargin(\example, y) \geq \epsilon - \marginthresh
    \quad \Rightarrow \quad \\
    \forall \advexample \in \mathcal{B}_\epsilon (\example) \; \baselineclf(\advexample) = y \vee \baselineclf(\advexample) = \abstainclass \\
\end{split}
\end{equation*}

Assume that $\signmargin(\example, y) \geq \epsilon - \marginthresh$. Let $\advexample \in \mathcal{B}_\epsilon (\example)$, i.e. $\|\example - \advexample\| < \epsilon$. By lemma \ref{lemma:margin_ineq},
$$|\signmargin(\example, y) - \signmargin(\advexample, y)| \leq \|\example - \advexample\| < \epsilon$$
Say $\signmargin(\example, y) \geq \signmargin(\advexample, y)$; then
$$\epsilon > \signmargin(\example, y) - \signmargin(\advexample, y) \geq \epsilon - \marginthresh - \signmargin(\advexample, y)$$
$$0 > - \marginthresh - \signmargin(\advexample, y)$$
$$\signmargin(\advexample, y) > -\marginthresh$$
Now say $\signmargin(\example, y) < \signmargin(\advexample, y)$; then
$$\signmargin(\advexample, y) > \signmargin(\example, y) \geq \epsilon - \marginthresh  \geq -\marginthresh$$
So in either case we have $\signmargin(\advexample, y) > -\marginthresh$. If $\signmargin(\advexample, y) \leq \marginthresh$, then $|\signmargin(\advexample, y)| \leq \marginthresh$ so $\baselineclf(\advexample) = \abstainclass$. Otherwise, if $\signmargin(\advexample, y) > \marginthresh \geq 0$, then $\baselineclf(\advexample) = \baseclf(\advexample) = y$. So
$$\forall \advexample \in \mathcal{B}_\epsilon (\example) \quad f(\advexample) = y \vee f(\advexample) = \abstainclass$$
\end{proof}

\section{Experimental Setup}
\label{experimental_setup}

We implement the baseline and \algac using PyTorch \citep{paszke_automatic_2017}. All training uses the Adam optimizer \citep{kingma_adam:_2014}. We preprocess images before classification by standardizing them based on the mean and standard deviation of each channel for all images in the dataset. The MNIST dataset can be obtained from \url{http://yann.lecun.com/exdb/mnist/} and the CIFAR-10 dataset can be obtained from \url{https://www.cs.toronto.edu/~kriz/cifar.html}.

\subsection{Network Architectures}
\label{network_arch}

For MNIST, we use train a 6-layer CNN architecture described in Table \ref{tab:mnist_arch}. For CIFAR-10, we train a WideResNet-28-5 \citep{zagoruyko_wide_2016}. We experimented with training a WideResNet-28-10, as was used in the original paper, but found that WideResNet-28-5 gave close the same performance with greatly reduced training and inference time. We use softplus activations for all networks, following \citep{qin_adversarial_2019}; we find that this reduces gradient masking, since the learned classifier functions are smooth.

\begin{table}[]
    \centering
    \begin{tabular}{ccccc}
        \toprule
        \bf Type & \bf Kernel & \bf Stride & \bf Output & \bf Activation \\
        \midrule
        Input & --- & --- & $28 \times 28 \times 1$ & --- \\
        Conv & $5 \times 5$ & $1 \times 1$ & $24 \times 24 \times 32$ & Softplus \\
        Conv & $5 \times 5$ & $1 \times 1$ & $20 \times 20 \times 32$ & Softplus \\
        MaxPool & $2 \times 2$ & $2 \times 2$ & $10 \times 10 \times 32$ & --- \\
        Conv & $3 \times 3$ & $1 \times 1$ & $8 \times 8 \times 64$ & Softplus \\
        Conv & $3 \times 3$ & $1 \times 1$ & $6 \times 6 \times 64$ & Softplus \\
        MaxPool & $2 \times 2$ & $2 \times 2$ & $3 \times 3 \times 64$ & --- \\
        Linear & --- & --- & 256 & Softplus \\
        Linear & --- & --- & 10 & Linear \\
        \bottomrule
    \end{tabular}
    \vspace{6pt}
    \caption{The CNN architecture used for MNIST experiments.}
    \label{tab:mnist_arch}
\end{table}

\subsection{Hyperparameters}
\label{hyperparameters}

All the hyperparameters used in experiments on MNIST and CIFAR-10 can be found in Table \ref{tab:hyperparameters}. We use identical hyperparameters for training the baseline and the proposed method, \algname.

\begin{table*}[]
    \centering
    \begin{tabular}{ccc}
        \toprule
        \bf Parameter & \bf MNIST & \bf CIFAR-10 \\
        \midrule
        Architecture & (see Table \ref{tab:mnist_arch}) & WideResNet-28-5 \\
        Num. parameters & 232,170 & 9,137,690 \\
        \midrule
        Optimizer & Adam & Adam \\
        Batch size & 100 & 50 \\
        Training epochs & 40 & 60 \\
        Learning rate & 0.001, 0.0001 & 0.001 \\
        Learning rate drop epoch & 30 & --- \\
        \midrule
        $\lambda$ & various & various \\
        $\eta$ & 0.02 & 0.02 \\
        \midrule
        Attack radius $\epsilon$ & 0.3 & 8/255 \\
        PGD iterations $T$ (train) & 10+ & 10 \\
        PGD loss (train) & $\left\{\ell_\text{abstain}, \ell_\text{sum}, \ell_\text{interp}\right\}$ & $\left\{\ell_\text{abstain}, \ell_\text{interp}\right\}$ \\
        PGD step size (train) & depends & depends \\
        \midrule
        PGD iterations $T$ (test) & 100 & 100 \\
        PGD step size (test) & $2 \epsilon / T = 0.006$ & $2 \epsilon / T \approx 0.0006$ \\
        DeepFool iterations (test) & 50 & 50 \\
        \bottomrule
    \end{tabular}
    \vspace{6pt}
    \caption{Hyperparameters used in the MNIST and CIFAR-10 experiments.}
    \label{tab:hyperparameters}
\end{table*}

\subsection{Attacks During Training}
\label{training_attacks_details}

During training, we use multiple variations of PGD to generate adversarial examples, as our experiments found that training against a single variation led to vulnerabilities against other variations. In particular, we find that $\ell_\text{abstain}$ and $\ell_\text{sum}$ tend to find adversarial examples in small gaps between the correct class and the abstain class. On the other hand, $\ell_\text{interp}$ and $\ell_\text{split}$ tend to find adversarial examples on the opposite side of the abstain class from the correct class.

For MNIST, we train against PGD using the $\ell_\text{abstain}$, $\ell_\text{sum}$, and $\ell_\text{interp}$ loss functions. PGD with $\ell_\text{interp}$ is always run for $T = 10$ iterations with step size $2 \epsilon / T = 0.06$. However, the other two loss functions are run until the loss converges; that is, until
$$
\left| \ell (f, \advexample^{(t)}, y) - \ell (f, \advexample^{(t+1)}, y) \right| < 0.1
$$
The step size used for these attacks is $\epsilon / (t + 5)$; that is, it harmonically decays with the attack step. We find that this combination of decaying steps size and iterating until convergence is necessary to prevent the network from masking its gradient during training.

For CIFAR-10, we train against PGD using the $\ell_\text{abstain}$ and $\ell_\text{interp}$ losses. On this dataset, PGD is always run for exactly $T = 10$ iterations. PGD with $\ell_\text{interp}$ uses step size $2 \epsilon / T = 0.006$. PGD with $\ell_\text{abstain}$ uses an exponentially decaying step size $\nicefrac{1}{2} \epsilon (0.8)^t$. See Figure \ref{fig:pgd_fixes} for the motivation for decaying the step size during the attack.

We do not train against DeepFool, leaving it as a previously unseen adversary to test against.

\section{Additional MNIST Results}
\label{additional_mnist}

Full results for our evaluation of the baseline and \algac on MNIST are shown in Table \ref{tab:error_mnist}.

\begin{table*}[t]
    \centering
    \begin{tabular}{lll|rrr|r|rrrrrrr}
        \toprule
        & & & \multicolumn{3}{c|}{\bf Natural Inputs} & \bf Noise & \multicolumn{7}{c}{\bf Adversarial Error} \\
        \bf Defense &  &  & Acc. & Abs. & Inc. & Abs. & $\ell_\text{abstain}$ & $\ell_\text{sum}$ & $\ell_\text{interp}$ & $\ell_\text{switch}$ & DF & DF-Abs & Any \\
        \midrule
        \bf Baseline & \multicolumn{2}{l|}{$\marginthresh = 0$} & 98.4 & 0.0 & 1.6 & 0.0 & --- & --- & --- & --- & --- & --- & 8.0 \\
        \bf Baseline & \multicolumn{2}{l|}{$\marginthresh = 0.1$} & 96.7 & 2.6 & 0.7 & 90.2 & --- & --- & --- & --- & --- & --- & 4.2 \\
        \bf Baseline & \multicolumn{2}{l|}{$\marginthresh = 0.2$} & 94.6 & 5.0 & 0.4 & 98.3 & --- & --- & --- & --- & --- & --- & 0.6 \\
        \bf Baseline & \multicolumn{2}{l|}{$\marginthresh = 0.3$} & 92.1 & 7.7 & 0.0 & 96.2 & --- & --- & --- & --- & --- & --- & 0.0 \\
        \midrule
        \bf \algac & $\ell^{(1)}$ & $\lambda = 4$ & 98.7 & 1.0 & 0.2 & 100.0 & 3.5 & 0.5 & 0.0 & 0.0 & 4.3 & 4.3 & 4.8 \\
        \bf \algac & $\ell^{(1)}$ & $\lambda = 8$ & 98.4 & 1.4 & 0.2 & 100.0 & 2.9 & 0.3 & 0.0 & 0.0 & 3.4 & 3.4 & 3.8 \\
        \bf \algac & $\ell^{(1)}$ & $\lambda = 16$ & 98.7 & 1.0 & 0.2 & 100.0 & 3.5 & 0.5 & 0.0 & 0.0 & 4.3 & 4.3 & 4.8 \\
        \bf \algac & $\ell^{(2)}$ & $\lambda = 1$ & 99.3 & 0.2 & 0.5 & 100.0 & 4.1 & 4.6 & 0.0 & 0.0 & 4.6 & 4.3 & 6.4 \\
        \bf \algac & $\ell^{(2)}$ & $\lambda = 2$ & 99.2 & 0.2 & 0.5 & 100.0 & 4.8 & 0.7 & 0.0 & 0.0 & 5.3 & 4.9 & 6.7 \\
        \bf \algac & $\ell^{(2)}$ & $\lambda = 4$ & 97.0 & 2.8 & 0.2 & 100.0 & 2.2 & 2.0 & 0.0 & 0.0 & 2.9 & 3.0 & 3.4 \\
        \bottomrule
    \end{tabular}
    \vspace{6pt}
    \caption{Identical evaluation to Table \ref{tab:error_cifar} but for MNIST.}
    \label{tab:error_mnist}
\end{table*}

\section{DeepFool Details}
\label{deepfool_details}

In Section \ref{attacks_deepfool}, we describe two variations of DeepFool \citet{moosavi-dezfooli_deepfool:_2016} that can be used to attack an abstaining classifier. Here, we give a more precise description of the original DeepFool algorithm and our variations.

\subsection{Original DeepFool algorithm}

Recall that for the classifier $\clf(\example)$ we denote the logit (activation before the softmax layer) for a class $i$ as $z_i(\example)$. Let $\hat{z}_{i,j}(\example) = z_i(\example) - z_j(\example)$, i.e. $\hat{z}_{i,j}(\example) = 0$ if $\example$ is on the decision boundary between the $i$th and $j$th classes.

DeepFool is an iterative attack; the attack starts with a natural example $\example = \advexample^{(0)}$ and at each step $t$ updates $\advexample^{(t)}$ to $\advexample^{(t+1)}$. The attack stops when misclassification is achieved, i.e. $\clf(\advexample^{(t)}) \notin \{y, \abstainclass\}$, or after a maximum number of steps $T$.

At each step, the attack considers every incorrect class $i \neq y$. DeepFool first calculates the nearest incorrect class under a linear approximation of the decision boundaries as
\begin{equation*}
    i^* = \argmin_{i \neq y}
    \frac{\left|\hat{z}_{y,i}(\advexample^{(t)})\right|}{\left\| \nabla_{\advexample} \hat{z}_{y,i}(\advexample^{(t)}) \right\|_1}
\end{equation*}
Then, DeepFool updates the current adversarial example with the minimal perturbation needed to reach this class under the linear approximation:
\begin{equation*}
    \advexample^{(t + 1)} \leftarrow
    \advexample^{(t)} +
    \frac{\hat{z}_{y,i}(\advexample^{(t)})}{\left\| \nabla_{\advexample} \hat{z}_{y,i}(\advexample^{(t)}) \right\|_1}
    \sign \big( \nabla_{\advexample} \hat{z}_{y,i}(\advexample^{(t)}) \big)
\end{equation*}
In the abstain setting, we make three small adjustments. First, we modify the function $\hat{z}_{y,i} (\advexample^{(t)})$:
\begin{equation*}
    \hat{z}_{y,i} (\advexample^{(t)}) =
    \max\left( z_y(\advexample^{(t)}), z_\abstainclass(\advexample^{(t)}) \right) - 
    z_i(\advexample^{(t)})
\end{equation*}
Second, when calculating $i^*$, we consider the minimum perturbation for $i \notin \{y, \abstainclass\}$ since our attack must fool both classifier and abstain class. Finally, the attack does not stop until the classifier outputs neither the correct class $y$ nor the abstain class $\abstainclass$.

\subsection{\dfname}

While the above modification of DeepFool for the abstain setting works in many cases, it only considers a single decision boundary it must cross at each step. However, this may cause the algorithm to sometimes cross the incorrect class's decision boundary but still remain in the abstain region (see Figure \ref{fig:deepfool_abstain}). Thus, we developed a second variation of DeepFool called \dfname, which we describe here.

Again define $\hat{z}_{i,j}(\example) = z_i(\example) - z_j(\example)$. At each step of \dfname, the attack considers every incorrect class $i \notin \{y, \abstainclass\}$. For each such $i$, it constructs a linear approximation of the boundaries between the incorrect class and both the correct class $y$ and abstain class $a$:
\begin{align*}
    L_{i, y}(\advexample^{(t)} + \delta) = \; &
    \hat{z}_{i,y}(\advexample^{(t)}) + \delta \, \nabla \hat{z}_{i,y}(\advexample^{(t)}) \\
    L_{i, a}(\advexample^{(t)} + \delta) = \; &
    \hat{z}_{i,a}(\advexample^{(t)}) + \delta \, \nabla \hat{z}_{i,a}(\advexample^{(t)})
\end{align*}
The algorithm attempts to update $\advexample^{(t)}$ to  $\advexample^{(t+1)}$ by adding a perturbation $\delta$ such that both these approximations are positive; that is, the updated adversarial example $\advexample^{(t+1)}$ is misclassified to be neither the correct class $y$ nor the abstain class $\abstainclass$. We would also like the $L_\infty$ norm of the perturbation update $\delta$ to be as small as possible to avoid moving the adversarial example outside of the acceptable perturbation neighborhood. This can be written as
\begin{align}
    \begin{split}
        \label{eq:deepfool_abstain_obj}
        \text{minimize} \quad
        & \| \delta \|_\infty \\
        \text{such that} \quad
        & L_{i, y}(\advexample^{(t)} + \delta) \geq 0 \\
        & L_{i, a}(\advexample^{(t)} + \delta) \geq 0
    \end{split}
\end{align}
We solve this convex optimization problem exactly for $\delta$ using the algorithm of Cadzow \citep{cadzow_finite_1973}, which has time complexity $O(d)$ where $d$ is the input space dimension. Then, the updated adversarial example $\advexample^{(t+1)}$ is given by
\begin{equation*}
    \advexample^{(t+1)} = \text{proj}\left(\advexample^{(t)} + \delta, \mathcal{B}_\epsilon(\example)\right)
\end{equation*}
That is, $\delta$ is added to the current adversarial example and then the result is projected into the ball of allowed adversarial examples.

See Algorithm \ref{alg:deepfool_abstain} for an overview of the \dfname attack. Figure \ref{fig:deepfool_abstain} shows an example of a single iteration of the attack.

\begin{algorithm}[t]
    \caption{The \dfname attack.}
    \label{alg:deepfool_abstain}
    \begin{algorithmic}
        \State $\advexample^{(0)} \gets \example$
        \For{$t = 0, \dots, T - 1$}
            \If {$\clf(\advexample^{(t)}) \notin \{y, \abstainclass\}$}
                \State $\advexample \gets \advexample^{(t)}$
                \State \textbf{return} $\advexample$
            \Else
                \State \textbf{solve} (\ref{eq:deepfool_abstain_obj}) \textbf{for} $\delta$
                \State $\advexample^{(t+1)} \gets \text{proj}\left(\advexample^{(t)} + \delta, \mathcal{B}_\epsilon(\example)\right)$
            \EndIf
        \EndFor
        \State $\advexample \gets \advexample^{(T)}$
        \State \textbf{return} $\advexample$
    \end{algorithmic}
\end{algorithm}

\section{Additional Comparisons to Related Work}
\label{related_work_additional}

Here we expand upon the descriptions of and comparisons to related work from Section \ref{related_work}.

Pang et al. \citet{pang_towards_2018} use an alternate training loss, reverse cross-entropy, to cluster inputs along a low-dimensional manifold in the final hidden layer's space of representations. They then detect as adversarial examples all inputs which produce a hidden layer representation far from this manifold by using kernel density estimation. They show that a detector-aware attack usually needs to apply macroscopic noise to generate an adversarial example that is not detected. One downside of their approach is that inference time complexity scales with the size of the training dataset. They do not report accuracy against this attack, so our method cannot be directly compared.

Liang et al. \citet{liang_detecting_2018}'s detection method classifies an input image both before and after applying denoising and spatial smoothing. If the two classes output are not the same, they detect the input as an adversarial example. However, their defense is still vulnerable to attack on 67\% of MNIST test samples; in contrast, our method \algac is vulnerable less than 5\% of the time (see Figure \ref{fig:error_plots} and Table \ref{tab:error_mnist}).

\citet{anonymous_adversarial_2019} begin with an already-trained classifier and train an additional detector for each class that attempts to distinguish adversarial examples from natural inputs. At inference time, the original classifier determines a class for an input; then, the corresponding detector is applied to determine if that input is adversarial. However, their defense remains worse than adversarial training when defending a detector-aware attack on MNIST; in contrast, our method strongly outperforms adversarial training in both accuracy and robustness (see Figure \ref{fig:error_plots} and Table \ref{tab:error_mnist}). Their method also requires training a separate network for every class, whereas ours uses a single network, improving training time.

\smallparagraph{Classification with an abstain option}
Some work has focused on training a classifier to abstain on certain inputs. Herbei and Wegkamp \citet{herbei_classification_2006} explore the Bayes-optimal abstention rule and its approximation in practice. Bartlett and Wegkamp \citet{bartlett_classification_2008} learn a classifier using a variation of the hinge loss modified for the abstain setting. However, little if any work has considered an abstain option in the context of adversarial robustness. While prior work has focused on adversarial detection, it has not explicitly described the motivation for detection and its resulting effect on the accuracy-robustness tradeoff.

\section{Calculating the Signed Margin Distribution}
\label{calculating_margin}

To use Theorem \ref{thm:margin}, we need to estimate the signed margin distribution for the base classifier $\baseclf (\cdot)$, since the theorem bounds are based on its CDF $F_\Gamma$. We estimate the distribution with two steps. First, we attack the base classifier $\baseclf (\cdot)$ with a number of adversarial attacks at varius radii. As stated in (\ref{eq:attack_bounds_margin}), an attack $\alpha_r$ succeeding at a given radius $r$ indicates that the signed margin is less than that radius. Thus, if the attack is strong enough, we can approximate the CDF of $\Gamma$:

\begin{equation*}
    F_\Gamma(r)
    = \mathop{\textbf{P}}_{(\example, y) \sim \mathcal{D}} (\signmargin(\example) \leq r)
    \approx \mathop{\textbf{P}}_{(\example, y) \sim \mathcal{D}} (\baseclf(\alpha_r(\example, y)) \neq y)
\end{equation*}

That is, for a given radius $r$ the value of $F_\Gamma(r)$ is the proportion of attacks that succeed at that radius.

Once we have approximated $F_\Gamma(r)$ for several values of $r$, we fit a normal CDF to the resulting values and find that it is a good fit for small positive values of $r$. Since $F_\Gamma$ is usually evaluated at small positive values in theorem 1, we use this normal approximation of $\Gamma$ to calculate the bounds shown in Figure \ref{fig:error_plots}.

\begin{figure}
    \centering
    \input{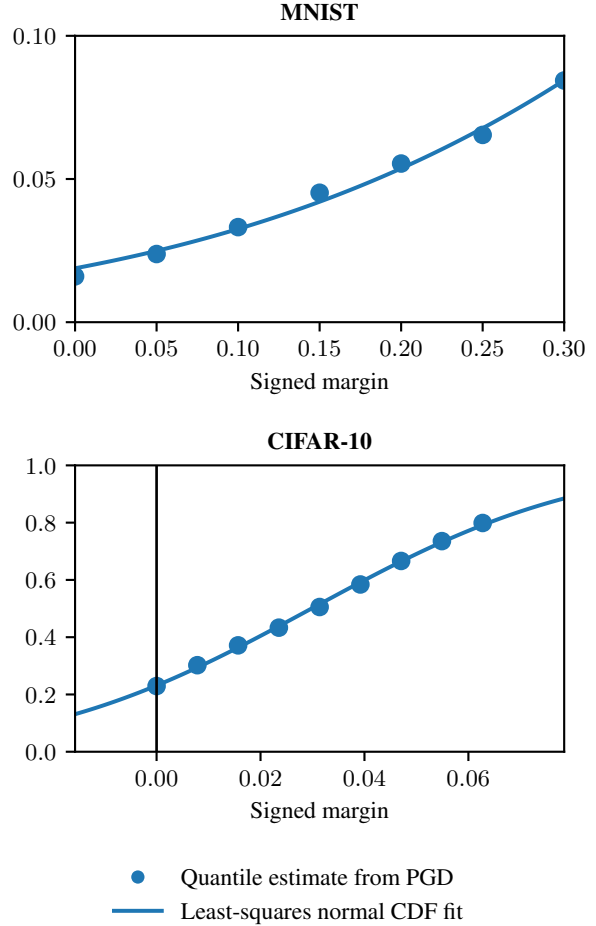}
    \caption{We approximate the distribution $\Gamma$ of signed margins for the adversarially trained base classifier $\baseclf(\cdot)$ by a normal distribution. First, quantiles of the distribution are estimated via an adversarial attack; then, a normal CDF is fit to those quantiles.}
    \label{fig:margin_cdf}
\end{figure}

For the adversarially trained classifier on MNIST, we find that $\Gamma \sim \mathcal{N}(0.887, 0.427^2)$ is a good approximation. For the adversarially trained classifier on CIFAR-10, we find that $\Gamma \sim \mathcal{N}(0.0298, 0.0406^2)$ is a good approximation. The fit of these distributions to the experimental data is shown in Figure \ref{fig:margin_cdf}. 

\end{document}